\documentclass[lettersize,journal]{IEEEtran}
\usepackage{amsmath,amsfonts}
\usepackage{algorithmic}
\usepackage{algorithm}
\usepackage{array}
\usepackage[caption=false,font=normalsize,labelfont=sf,textfont=sf]{subfig}
\usepackage{textcomp}
\usepackage{stfloats}
\usepackage{url}
\usepackage{verbatim}
\usepackage{graphicx}
\usepackage{cite}
\usepackage{tcolorbox}
\usepackage{balance}
\usepackage{booktabs}
\usepackage{arydshln}

\usepackage{pifont}
\usepackage{wrapfig}
\usepackage{svg} 
\usepackage{colortbl}  
\usepackage{amssymb}

\usepackage{multirow}
\usepackage{indentfirst}
\usepackage{threeparttable}
\usepackage{makecell}
\usepackage{tabularx}
\usepackage[table]{xcolor}
\newcommand{\blue}[1]{\textbf{\textcolor{mblue}{#1}}}
\newcommand{\red}[1]{\textbf{\textcolor{mred}{#1}}}
\definecolor{mblue}{RGB}{0, 77, 128}

\definecolor{mblue}{RGB}{0, 77, 128}
\definecolor{mred}{RGB}{192,0, 0}
\definecolor{darkgreen}{rgb}{0.0, 0.5, 0.0}
\definecolor{mycolor_blue}{HTML}{E7EFFA}
\definecolor{mycolor_gray}{HTML}{ECECEC}
\newcommand{\cmark}{\textcolor{darkgreen}{\ding{51}}}  
\newcommand{\xmark}{\textcolor{red}{\ding{55}}}        

\begin{document}

\title{3DGSI-Assessor: A Large-Scale Dataset and An LMM-based Method for 3D Gaussian Splatting Image Quality Assessment}

\author{Yuke Xing, Jiarui Wang, William Gordon, Zhu Li,~\IEEEmembership{Senior Member,~IEEE,}Guangtao Zhai,~\IEEEmembership{Fellow,~IEEE,} and Yiling Xu,~\IEEEmembership{Member,~IEEE}
\thanks{Yuke Xing, Jiarui Wang, Guangtao Zhai and Yiling Xu are with the School of Information Science and Electronic Engineering,
Shanghai Jiao Tong University, Shanghai 200240, China (e-mail: {xingyuke-v, wangjiarui, zhaiguangtao, yl.xu}@sjtu.edu.cn).}
\thanks{Zhu Li is with the School of Science and Engineering,
University of Missouri–Kansas City, Kansas City, MO 64110 USA (e-mail:lizhu@umkc.edu).}
\thanks{William Gordon is with the Basis Independent Silicon Valley, San Jose, CA 95134, USA (e-mail:williamg.research@gmail.com).}
\thanks{Corresponding author: Yiling Xu.}}



\maketitle

\begin{abstract}

3D Gaussian Splatting (3DGS) has become a dominant representation for real-time novel view synthesis (NVS), yet its storage footprint makes compression indispensable for practical deployment. 3DGS training and compression introduce representation-specific distortions such as floating artifacts and surface scattering, which conventional image quality assessment (IQA) metrics fail to capture. 
Moreover, the independent compression of geometric and color attributes may lead to decoupled dimension-specific distortions that must be diagnosed separately, yet existing metrics report only a single overall score. 
To address these gaps, we present 3DGS-IEval-15K+, a large-scale, multi-dimensional IQA dataset for compressed 3DGS, comprising 15,200 images from 10 diverse scenes, produced by 6 representative 3DGS algorithms at systematically designed compression levels and rendered from 20 strategically selected viewpoints spanning both training views and challenging novel views, annotated with 45,600 mean opinion scores (MOSs) across overall, geometry, and color quality. 
Based on 3DGS-IEval-15K+, we propose 3DGSI-Assessor, an all-in-one 3DGS IQA framework that integrates global semantic and dimension-specific local features within a large multimodal model (LMM), predicting all three dimensions in a single forward pass. 3DGSI-Assessor achieves state-of-the-art performance on 3DGS-IEval-15K+, and exhibits competitive generalization on other NVS benchmarks. Dataset and code will be released at \url{https://github.com/YukeXing/3DGSI-Assessor}.
\end{abstract}

\begin{IEEEkeywords}
3D Gaussian-splatting, image quality assessment, compression, large multimodal model, benchmark
\end{IEEEkeywords}
\section{Introduction}

\begin{figure}
    \centering
    \includegraphics[width=1\linewidth]{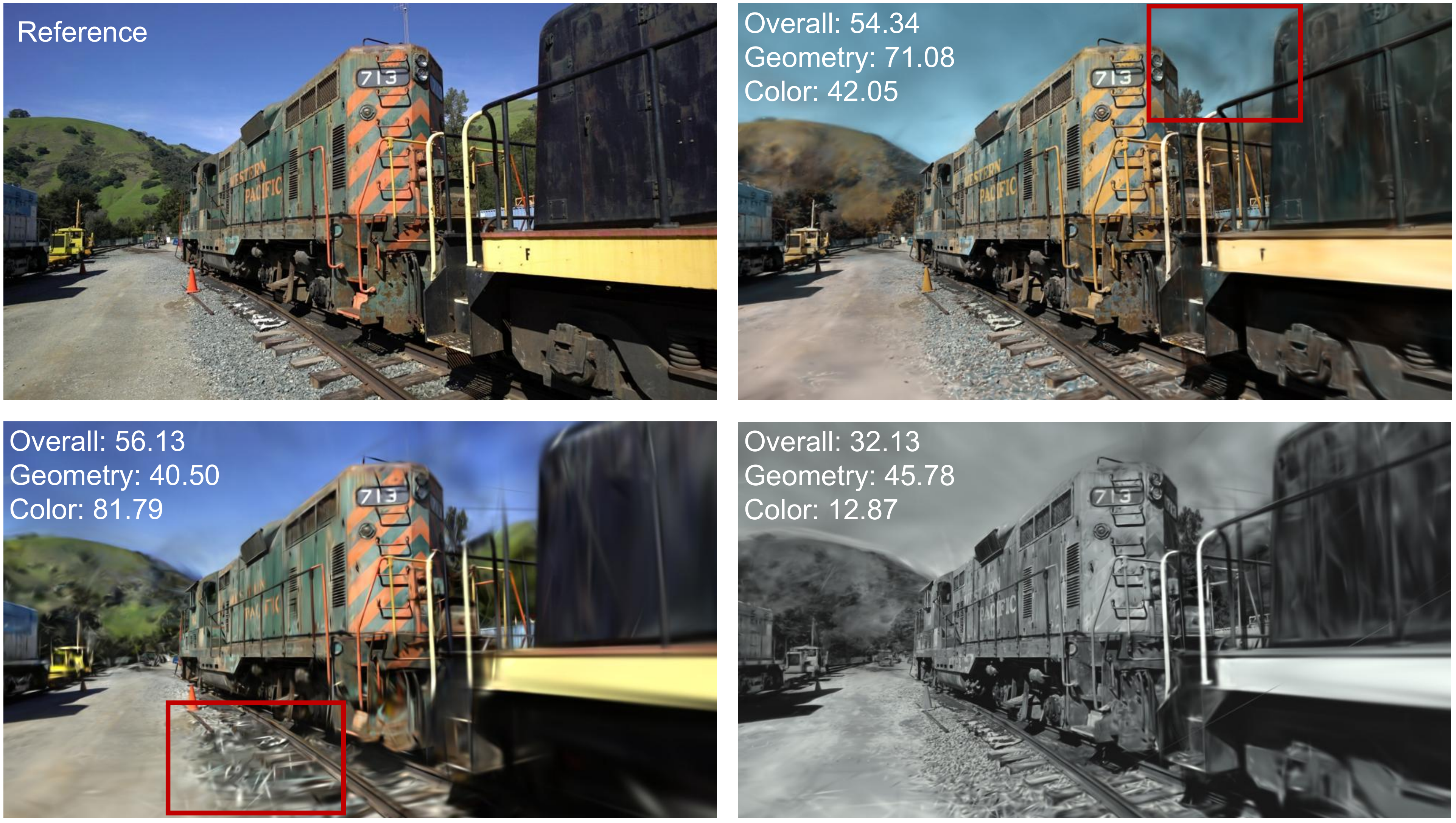}
    \vspace{-6mm}
    \caption{Illustration of multi-dimensional quality assessment for generative compressed 3DGS. Different compression methods and configurations result in distinct degradation patterns: the top-right image exhibits moderate geometry degradation with notable color distortion, the bottom-left shows notable geometry degradation while preserving color fidelity, and the bottom-right suffers from severe color degradation. Red boxes highlight characteristic 3DGS distortions including Gaussian ellipsoid artifacts and floating distortions.} 
    \label{fig:teasor1}
    \vspace{-4mm}
\end{figure}

\begin{figure*}[!t]
    \centering
    \vspace{-1mm}
    \includegraphics[width=1\linewidth]{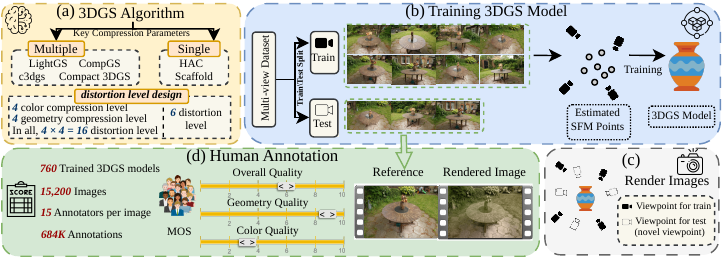} 
    \vspace{-5mm}
    \caption{Overview of the construction pipeline of 3DGS-IEval-15K+, designed for benchmarking 3DGS representations. (a) We select 6 mainstream compressed 3DGS algorithms, including 4 multi-compression-parameter methods designed with 4 geometry and color compression levels each (combined to produce 16 distinct distortion levels), and 2 single-compression-parameter methods designed with 6 distortion levels each. (Section~\ref{3DGS_Model_and_Bitrate_Point_Selection}) (b) We select 10 scenes from multi-view datasets, each containing numerous viewpoints. For each scene, we split the viewpoints into training/testing set and use the training set to reconstruct 3D scenes. (Section~\ref{Source_Content_Selection}) (c) For each scene, we select 10 representative viewpoints from the training set and 10 challenging viewpoints from the testing set as defined in (b). Reconstructed 3D scenes are then rendered from these 20 viewpoints. (Section~\ref{3D_Viewpoint_Selection}) (d) With collecting 684K human annotations across three evaluation dimensions: overall quality, geometry quality, and color quality, we create a large-scale, multi-dimensional 3DGS image quality evaluation dataset (3DGS-IEval-15K+) with labeled MOS scores. (Section~\ref{Subjective_Experiment_and_Data_Processing})}
    \label{fig:teasor2}
    \vspace{-2mm}
\end{figure*}

\IEEEPARstart{N}{ovel} View Synthesis (NVS) has made significant advancement in computer vision, enabling photorealistic renderings from arbitrary viewpoints using only sparse input observations.
Recently, NVS has evolved from Neural Radiance Fields (NeRF) \cite{nerf} to 3D Gaussian Splatting (3DGS) \cite{3DGS}, which enables real-time, high-fidelity rendering. 
However, this explicit representation demands significant storage resources, substantially hindering practical deployment.
In response, recent state-of-the-art (SOTA) 3DGS algorithms \cite{Scaffold,HAC,LightGS,CompGS,c3dgs,Compact-3DGS,eagles} increasingly incorporate dedicated compression modules.
This trend creates a critical need for specialized Image Quality Assessment (IQA) metrics to faithfully diagnose the quality degradations introduced by 3DGS training and compression \cite{3DGS,Scaffold,HAC,LightGS,CompGS,c3dgs,Compact-3DGS,eagles,TVCG-VolSegGS}.

Evaluating 3DGS images poses unique challenges due to specific distortion mechanisms introduced by its generation and compression pipelines. 
First, the unique representation of 3DGS introduces characteristic artifacts\cite{ENeRF-QA,NVS-SQA,3DGS-VBench}, such as floating artifacts, surface scattering \cite{ENeRF-QA}, and color inconsistencies caused by training view overfitting, especially under sparse input conditions \cite{sparse-view-3dgs}. Second, 3DGS explicitly constructs scenes using Gaussian primitives with learnable parameters that can be categorized into two types: geometric attributes (position, covariance, and opacity) defining spatial structure, and color attributes (spherical harmonic coefficients) determining appearance.  
In representative 3DGS generative compression methods\cite{LightGS,CompGS,c3dgs,Compact-3DGS,eagles}, these geometric and color attributes are controlled by independent compression parameters, each requiring separate tuning for optimal quality-size trade-offs.
This leads to a unique distortion paradigm where geometric and color degradations occur independently and must be assessed separately, as shown in Figure \ref{fig:teasor1}.

For evaluation of NVS, the most widely used metrics, such as PSNR, SSIM\cite{PSNR}, which measure pixel-wise fidelity rather than perceptual quality, have been proven inadequate for assessing the perceptual quality of 3DGS images in empirical studies\cite{NVS-SQA,3DGS-VBench,ENeRF-QA,3DGS-IEval-15K}. 
Moreover, these metrics cannot decouple dimension-specific distortions, hindering their ability to diagnose specific degradations.
Consequently, there is an urgent need for specialized 3DGS-tailored IQA metrics capable of capturing characteristic distortions and providing accurate, multi-dimensional assessments across overall, geometric, and color quality. 
Beyond evaluation, IQA metrics are also used as visual reward signals to guide 3DGS training and compression\cite{LightGS,CompGS,c3dgs,Compact-3DGS,eagles}, a role currently served by conventional metrics such as PSNR and SSIM. A more accurate, dimension-specific metric therefore can provide better feedback, \textit{e.g.}, indicating whether the geometric or the color compression parameter should be re-tuned to recover quality at a given storage budget. 



To develop specialized evaluation metrics better aligning with human judgment, several NVS subjective quality assessment datasets with human annotations\cite{ENeRF-QA,NeRF-VSQA,GSC-QA,3DGS-VBench,NeRF-QA,FFV,GS-QA,NVS-QA} have been established, as shown in Table \ref{tab:relate}.
However, these benchmarks exhibit several critical limitations: 
(1) \textbf{Severely constrained scale.} Most datasets contain fewer than 200 samples due to high training costs, with only one exceeding 500\cite{3DGS-VBench}, which is insufficient for training robust objective models. 
(2) \textbf{Limited distortion diversity.} Despite compression being a research priority, many datasets \cite{NeRF-QA,NeRF-VSQA,FFV,GS-QA,NVS-QA} lack systematic compression parameter designs, resulting in inadequate distortion diversity for generalizable metric training. 
(3) \textbf{Neglect of 3DGS IQA.} Although video quality assessment (VQA) \cite{NeRF-VSQA,ENeRF-QA,GS-QA,3DGS-VBench} simulates continuous view traversal, per-view IQA is crucial in 3DGS, since reconstruction quality varies strongly with viewpoint; attributing degradations to individual rendered views rather than averaging them along a camera trajectory is essential for analyzing the view-dependent quality variations characteristic of 3DGS.
(4) \textbf{Lack of multi-dimensional annotations.}
Existing datasets provide only coarse overall quality scores. Without multi-dimensional labels for overall, geometry, and color dimensions, they cannot support the development of IQA algorithms capable of providing dimension-specific diagnostic feedback.

To address these limitations, we present \textbf{3DGS-IEval-15K+}, the first large-scale, multi-dimensional IQA dataset for 3DGS. It comprises 15,200 images derived from 10 diverse scenes and generated by 6 representative 3DGS compression algorithms with different compression levels leading to various distortion effects. As illustrated in Figure \ref{fig:teasor2}, we collect about 684K human annotations, yielding 45,600 Mean Opinion Scores (MOSs) across three dimensions: overall, geometry, and color quality. 
Based on 3DGS-IEval-15K+, we propose \textbf{3DGSI-Assessor}, a large multimodal model (LMM)-based \textbf{all-in-one} multi-dimensional IQA metric for 3DGS. Specifically, 3DGSI-Assessor adopts a hierarchical visual encoding strategy that integrates global semantic representations with local dimension-specific geometry and color features, which are then fused through an LMM \cite{InternVLmodel}.  
We adopt a two-stage training strategy with low-rank adaptation (LoRA)\cite{LORA} for more efficient feature refining: first aligning visual features with linguistic space via instruction tuning \cite{instruction-tuning}, then specializing the LMM for multi-dimensional score regression.
Extensive experiments demonstrate that 3DGSI-Assessor achieves SOTA performance on 3DGS-IEval-15K+ and exhibits competitive zero-shot generalization ability across other NVS benchmarks.

The main contributions of this work are summarized as follows:
\begin{itemize}
\item{We construct 3DGS-IEval-15K+, the first large-scale IQA dataset for 3DGS that, to our knowledge, is the first to provide dimension-separated geometry and color annotations. It comprises 15,200 images from 10 diverse scenes, generated by 6 representative 3DGS compression algorithms at various compression levels. Compared with the preliminary conference version 3DGS-IEval-15K\cite{3DGS-IEval-15K}, it extends the annotation scheme from a single overall dimension to three fine-grained dimensions, tripling the annotation volume to 45,600 MOSs and enabling dimension-specific diagnostic feedback.}
\item{We propose 3DGSI-Assessor, a unified LMM-based metric for 3DGS IQA that couples global semantic representations with dimension-specific local features, predicting all three dimensions in a single forward pass.}
\item{Extensive experimental results on 3DGS-IEval-15K+ with more than 30 existing IQA metrics across all three dimensions and other NVS benchmarks validate the SOTA performance and competitive generalization ability of 3DGSI-Assessor.}
\end{itemize}

\section {Related Work}

\begin{table*}[!]
\centering
\vspace{-5mm}
\caption{Summary for existing NVS quality evaluation datasets, with ``Syn'' and ``Real'' representing synthetic and real scenes. }
\vspace{-1mm}
\renewcommand\arraystretch{0.85}
\label{tab:relate}
\resizebox{1\textwidth}{!}{\begin{tabular}{ccccccccccccc}
\hline
Dataset & \multirow{2}{*}{Name} & \multirow{2}{*}{Year}  
&\multicolumn{2}{c}{Scenes} & 
\multicolumn{2}{c}{NVS Models} 
& Render
& Distortion 
&\multirow{2}{*}{Annotation}
& Number of & Number of &  \multirow{2}{*}{MOSs/JODs}
\\

 Type & & & Syn & Real & NeRF & 3DGS & Mode & Level design & & NeRF/3DGS &videos/imgs \\
 \hline

\multirow{7}{*}{VQA} & NeRF-QA\cite{NeRF-QA}  & 2023 & 4 & 4 & 7 & - &$360^{\circ}$ & \xmark & DMOS & 48 & 48 &48\\
& NeRF-VSQA\cite{NeRF-VSQA} & 2024 & 8 & 8  & 7 & - & $360^{\circ}$+Front & \xmark & DMOS & 88 & 88 & 88 \\
& FFV\cite{FFV} & 2024 & - & 22 & 8 & - & \textit{Front} & \xmark & Pairs & 220 & 220 & 144 \\
& ENeRF-QA\cite{ENeRF-QA} & 2024 & 22 & - &  4 & - & $360^{\circ}$ & \cmark & MOS & 440 & 440 & 440 \\
& GSC-QA\cite{GSC-QA} & 2024 & 9 & 6 & - & 1 & $360^{\circ}$ & \cmark & MOS & 120 & 120 & 120 \\
& GS-QA \cite{GS-QA}& 2025 & - & 8 & - & 7 & $360^{\circ}$ + Front & \xmark & DMOS & 64 & 64 & 64 \\
& NVS-QA (video)\cite{NVS-QA} & 2025 & - & 13 & 2 & 3 & $360^{\circ}$+Front & \xmark & MOS & 65 & 65 & 65 \\
& 3DGS-VBench\cite{3DGS-VBench} & 2025 & - & 11 & - & 6 & $360^{\circ}$ & \cmark & MOS & 660 & 660 & 660 \\
\cdashline{1-13}[7pt/5pt]  
\multirow{2}{*}{IQA} & NVS-QA (image)\cite{NVS-QA} & 2025 & - & 13 & 2 & 3 & 1 viewpoints& \xmark & MOS & 65 & 65 & 65 \\
& \cellcolor{gray!20}\textbf{3DGS-IEval-15K+(Ours)} 
& \cellcolor{gray!20}- 
& \cellcolor{gray!20}-
& \cellcolor{gray!20}\textbf{10} 
& \cellcolor{gray!20}- 
& \cellcolor{gray!20}\textbf{6} 
& \cellcolor{gray!20}\textbf{20 viewpoints}
& \cellcolor{gray!20}\cmark 
& \cellcolor{gray!20}\textbf{MOS} 
& \cellcolor{gray!20}\textbf{760} & \cellcolor{gray!20}\textbf{15,200} 
&\cellcolor{gray!20}\textbf{45,600}\\
\hline
\end{tabular}}
\end{table*}

\begin{figure*}[!t]
    \centering
    \vspace{-1mm}
    \includegraphics[width=1\linewidth]{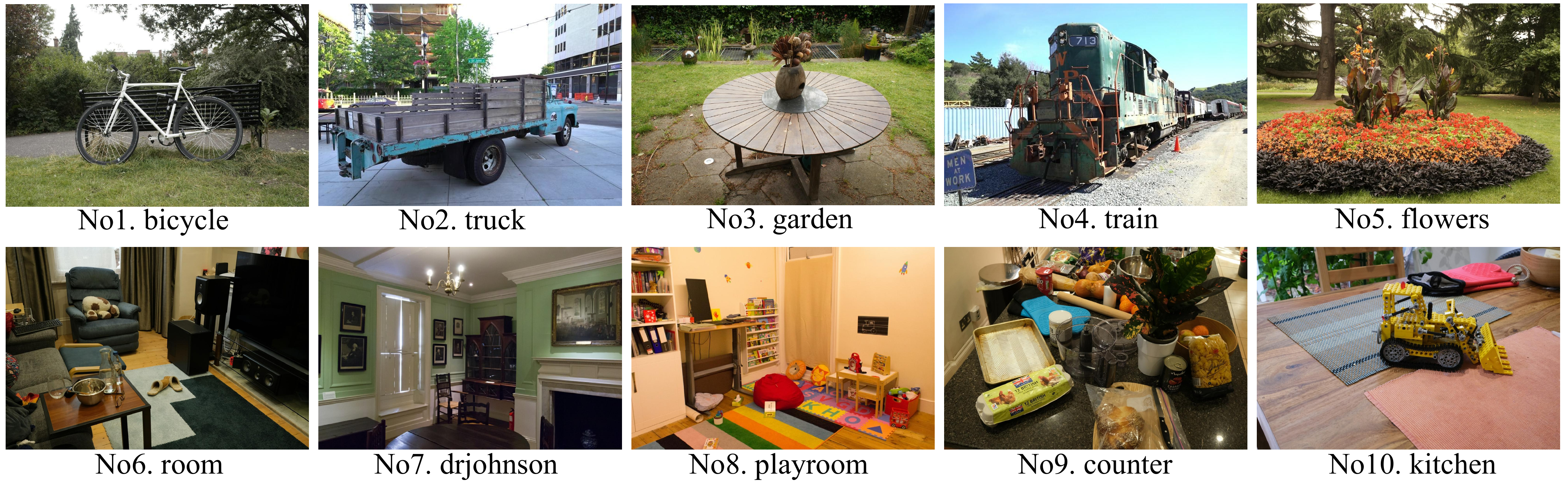}
    \vspace{-7mm}
    \caption{10 selected source content in 3DGS-IEval-15K+: scenes 1-5 depict outdoor scenes, while scenes 6-10 depict indoor scenes.} 
    \label{figure:model}
    \vspace{-2mm}
\end{figure*}

\subsection{Quality Assessment Databases for NVS }
The evolution of quality assessment approaches for NVS has closely tracked technological developments in the field, exhibiting a clear transition from NeRF-based evaluation to 3DGS-focused methodologies, as documented in Table~\ref{tab:relate}. Early benchmarking efforts focused on NeRF datasets: NeRF-QA \cite{NeRF-QA} and NeRF-VSQA \cite{NeRF-VSQA} established initial assessment frameworks with 48 and 88 video samples respectively, whereas FFV \cite{FFV} introduced pairwise comparison protocols using 220 samples. ENeRF-QA \cite{ENeRF-QA} further advanced the field by systematically constructing distortions via NeRF compression techniques across 440 samples, concurrently characterizing nine distortion types unique to NeRF technology. The emergence of 3DGS as the preferred NVS technology catalyzed a corresponding transformation in evaluation benchmarks: GSC-QA \cite{GSC-QA} explored compression impacts of their 3DGS compression method using 120 samples; GS-QA \cite{GS-QA} conducted comparative evaluation between 3DGS and NeRF across 64 samples; and NVS-QA \cite{NVS-QA} combined video and image assessment, each incorporating 65 samples. Most recently, 3DGS-VBench \cite{3DGS-VBench} introduced a large-scale VQA benchmark containing 660 compressed 3DGS video sequences across 6 3DGS compression algorithms.
However, existing datasets predominantly focus on VQA while neglecting IQA, provide only overall quality scores without multi-dimensional assessments, and remain limited in scale and distortion diversity. 
To address these gaps, we present 3DGS-IEval-15K+, a large-scale IQA benchmark expressly designed for 3DGS assessment, featuring 15,200 samples with multi-dimensional annotations.

\subsection{Evaluation Metrics for NVS}
Various image quality assessment models have been introduced in the literature. Handcrafted metrics such as PSNR\cite{PSNR}, SSIM\cite{PSNR} and LPIPS\cite{LPIPS} are widely used for NVS evaluation, yet empirical studies\cite{NVS-SQA,3DGS-VBench,ENeRF-QA,3DGS-IEval-15K} confirm their shortcomings on 3DGS images.
LMMs\cite{LLaVA-1.5,Llava-one-vision,LLaVA-NeXT,DeepSeekVL,Qwen2.5-VL}, despite their advanced visual understanding, exhibit limited proficiency in fine-grained perceptual quality assessment and often struggle to deliver accurate scores for 3DGS-rendered images with distinctive distortions. 
Deep learning-based models\cite{CSVTlearningbasedQA,DBCNN,HYPERIQA,MUSIQ,CSVTlearningbasedPQA1} can learn diverse distortion patterns but cannot perform the multi-dimensional assessment crucial for 3DGS quality monitoring and compression optimization.
Beyond image-level metrics, a pioneering line of work has established dedicated objective quality assessment for neurally synthesized scenes. 
NVS-SQA\cite{NVS-SQA} learns quality representations via self-supervised contrastive learning, while NeRF-NQA\cite{NeRF-NQA} performs no-reference, scene-level assessment over COLMAP-sampled surface points. 
These methods are effective within their intended settings, but adopt an assessment paradigm different from ours. 
NVS-SQA\cite{NVS-SQA} is calibrated for within-scene relative ranking rather than cross-scene absolute scoring, and is thus not directly applicable when absolute quality must be compared across scenes and methods, as required for application-side quality monitoring. 
NeRF-NQA\cite{NeRF-NQA} produces a single score per reconstructed scene and requires as input a densely sampled view sequence with camera poses and a COLMAP sparse point cloud. Such multi-view, geometry-dependent inputs are unavailable in the per-image IQA setting targeted here as well as some practical scenarios, and are absent from many subjective 3DGS datasets, including ours. And its original formulation is scoped to front-facing NeRF scenes, explicitly leaving 360$^{\circ}$ content and 3DGS to future work. NeRF-NQA is therefore outside the scope of our per-image, multi-dimensional comparison.
Motivated by these observations, we propose an LMM-based full-reference IQA framework that leverages our large-scale annotated dataset to provide all-in-one, multi-dimensional 3DGS assessment.

\section{DATABASE CONSTRUCTION}

\begin{figure*}[!t]
    \centering
     \vspace{-5mm}
    \includegraphics[width=0.8\linewidth]{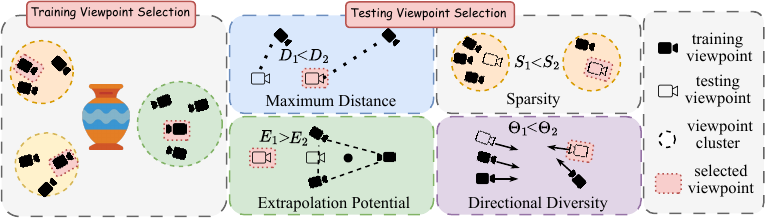} 
    \label{fig:train_test_selection}
    \vspace{-2mm}
    \caption{ Illustration of the proposed viewpoints selection strategy: training viewpoints are selected through feature-based $k$-means clustering, while testing viewpoints are chosen based on four criteria.}
    \label{fig:selection_combined}
    \vspace{-2mm}
\end{figure*}

\subsection{Source Content Selection}
\label{Source_Content_Selection}

Our dataset consists of 10 real-world scenes from three canonical multiview datasets in the NVS domain to ensure diverse visual characteristics. As demonstrated in Figure~\ref{figure:model}, the collection encompasses six scenes from \textit{Mip-NeRF 360}\cite{Mip-NeRF-360}, featuring three outdoor scenes: \textit{flowers} (1256~$\times$~828), \textit{bicycle} (1237~$\times$~822),  \textit{garden} (1297~$\times$~840), alongside three indoor scenes: \textit{kitchen} (1558~$\times$~1039), \textit{counter} (1558~$\times$~1038), \textit{room} (1557~$\times$~1038). Furthermore, two outdoor scenes are extracted from \textit{Tanks \& Temples}\cite{Tanks-Temples}: \textit{truck} (979~$\times$~546), \textit{train} (980~$\times$~545), while two indoor scenarios are obtained from \textit{Deep Blending}\cite{Deep-Blending}: \textit{drjohnson} (1332~$\times$~876), \textit{playroom} (1264~$\times$~832).
This selection ensures diversity and challenge for 3DGS training and quality assessment, including varying lighting conditions, reflective materials, diverse indoor and outdoor environments, intricate occlusion patterns, and natural textures with high-frequency details.
We maintain the native resolution settings from the original 3DGS work\cite{3DGS} to ensure consistency with prior research and facilitate fair cross-study communication.
For each scene in these multiview databases, the viewpoints are partitioned into disjoint training and testing set. The training set is utilized for 3DGS model training, while the testing set serves as reference to evaluate the quality of the corresponding novel views reconstructed by 3DGS.

\subsection{3D Viewpoint Selection}
\label{3D_Viewpoint_Selection}

Given the vast number of viewpoints per scene in multi-view datasets and the labor-intensive nature of human assessment, selecting a diagnostic subset of viewpoints is indispensable for constructing 3DGS subjective IQA datasets. To this end, we introduce a systematic strategy (Figure~\ref{fig:selection_combined}) that identifies $\mathbf{10}$ representative viewpoints from the training set and $\mathbf{10}$ challenging viewpoints from the testing set as outlined in Section~\ref{Source_Content_Selection}, facilitating robust assessment of view-dependent generalization and perceptual quality under diverse conditions.

\textbf{Training Viewpoint Selection:} we employ a feature-driven clustering approach to select training viewpoints, ensuring broad scene coverage while minimizing redundancy. Within each scene, every viewpoint from the training set is encoded by a feature vector $\mathbf{f} = [\mathbf{p}, \beta\mathbf{d}]$, where $\mathbf{p}$ denotes the position, $\mathbf{d}$ denotes the viewing direction, and $\beta = 0.3$ balances their relative importance. These features are then
normalized and partitioned using $k$-means clustering, yielding $\mathbf{10}$ distinct clusters. Within each cluster, the viewpoint nearest to the centroid is chosen, guaranteeing that the selected viewpoints optimally represent the distribution of available camera poses while preserving diversity.

\textbf{Testing Viewpoint Selection:} to stringently evaluate the generalization capability of NVS, we identify testing viewpoints that demonstrate maximal distributional divergence from the training set viewpoints. Each candidate is ranked using a composite score that combines multiple criteria. The composite score for a testing set viewpoint $j$ is expressed as:
\vspace{-5pt}
\begin{equation}
S_j = w_d D_j + w_s L_j + w_e E_j + w_\theta \Theta_j
\vspace{-5pt} 
\end{equation}
where $D_j$ denotes normalized distance metrics, $L_j$ characterizes local sparsity, $E_j$ specifies extrapolation requirements, $\Theta_j$ evaluates directional novelty, and $w$ factors denote the corresponding weights (we set all $w$ values to 0.25). The precise definitions of these four criteria are provided below:\\\
\textit{Maximum Distance}: preference for viewpoints maximally distant from any training viewpoints:
\vspace{-5pt} 
\begin{multline}
D_j = \frac{1}{2} \Bigg( \frac{\min_{i \in T} \|\mathbf{p}_j - \mathbf{p}_i\|}{\max_k \min_{i \in T} \|\mathbf{p}_k - \mathbf{p}_i\|} \\
+ \frac{\frac{1}{|T|}\sum_{i \in T} \|\mathbf{p}_j - \mathbf{p}_i\|}{\max_k \frac{1}{|T|}\sum_{i \in T} \|\mathbf{p}_k - \mathbf{p}_i\|} \Bigg)
\end{multline}
\textit{Local Sparsity}: prioritization of viewpoints in regions with low training viewpoint density:
\vspace{-1.2mm}
\begin{multline}
L_j = 1 - \frac{\rho_j - \min_k \rho_k}{\max_k \rho_k - \min_k \rho_k}, \\
\text{where} \quad \rho_j = \frac{1}{\frac{1}{K}\sum_{l=1}^K d_{j,l} + \epsilon}
\end{multline}
\textit{Extrapolation Potential}: identifying viewpoints outside the convex hull of training viewpoints, requiring extrapolation rather than interpolation:
\vspace{-4pt} 
\begin{equation}
E_j = \begin{cases}
1 & \text{if } \mathbf{p}_j \notin \text{ConvexHull}(\{\mathbf{p}_i\}_{i \in T}) \\
0 & \text{otherwise}
\end{cases}
\vspace{-4pt} 
\end{equation}
\textit{Directional Diversity}: prioritizing viewing directions substantially different from training viewpoints:
\vspace{-5pt} 

\begin{equation}
\Theta_j = \frac{\phi_j - \min_k \phi_k}{\max_k \phi_k - \min_k \phi_k},
\end{equation}
where $\phi_j = \max_{i \in T} \arccos(\mathbf{d}_j \cdot \mathbf{d}_i)$ represents the maximum angular deviation, $T$ denotes the indices of training set viewpoints, $p_j$ and $\mathbf{d}_j$ denote the position and viewing direction of viewpoint $j$, $k$ iterates over all testing set viewpoints, $d_{j,l}$ denotes the distance from viewpoint $j$ to its $l$-th nearest training set viewpoints, $K$ denotes the number of nearest training set viewpoints for viewpoint $j$ (set as $K=min(10, |T|)$), $\epsilon$ denotes a small constant to avoid division by zero. The composite scores are calculated for all testing set viewpoints and we identify the top $\mathbf{10}$ with the highest scores.

This targeted selection strategy offers more stringent evaluation than random sampling by emphasizing viewpoints requiring significant interpolation or extrapolation from training data, facilitating dependable evaluation of model generalization in challenging novel view synthesis scenarios.

\vspace{-2pt} 
\subsection{3DGS Model and Bitrate Point Selection}
\label{3DGS_Model_and_Bitrate_Point_Selection}

\begin{table*}[tbph]
\centering
\vspace{-5mm}
\renewcommand{\arraystretch}{1}
\caption{
4 CL designed separately for color-distortion-controlled compression parameters and geometry-distortion-controlled compression parameters for multi-compression-parameter methods, with 16 DL 3DGS models each generated through pairwise combinations of these parameters with various CL.} 
\vspace{-2mm}
\resizebox{1\linewidth}{!}{
\begin{tabular}{c!{\vrule}c:c!{\vrule}c:c!{\vrule}c:c!{\vrule}c:c}
    \toprule[1pt]
    \textbf{3DGS Model} & \multicolumn{2}{c!{\vrule}}{\textbf{LightGS}\cite{LightGS}} & \multicolumn{2}{c!{\vrule}}{\textbf{C3dGS}\cite{c3dgs}} & \multicolumn{2}{c!{\vrule}}{\textbf{Compact-3DGS}\cite{Compact-3DGS}} & \multicolumn{2}{c}{\textbf{CompGS}\cite{CompGS}} \\
    \midrule
    \textbf{Distortion Type} & \textbf{Color} & \textbf{Geometry} & \textbf{Color} & \textbf{Geometry} & \textbf{Color} & \textbf{Geometry} & \textbf{Color} & \textbf{Geometry} \\
    \midrule
    \textbf{Key} & vq\_ratio & \multirow{2}{*}{prune\_percents} & codebook\_size & codebook\_size & \multirow{2}{*}{hashmap} & codebook\_size & \multirow{2}{*}{codebook\_size} & \multirow{2}{*}{codebook\_size} \\
    \textbf{Parameters}& codebook\_size && importance\_include & importance\_include && rvq\_num &&\\
    \midrule
    \textbf{CL01} & (1, $2^1$) & 0.95 & (2, 0.6) & (1, 0.3) & 2 & ($2^2$, 1) & $2^1$ & 1 \\
    \textbf{CL02} & (1, $2^2$) & 0.90 & ($2^2$, 0.6) & ($2^2$, 0.3) & $2^2$ & ($2^4$, 1) & $2^2$ & 2 \\
    \textbf{CL03} & (1, $2^5$) & 0.85 & ($2^5$, 0.6) & ($2^4$, 0.3) & $2^9$ & ($2^2$, 6) & $2^3$ & $2^3$ \\
    \textbf{CL04} & (0.6, $2^{13}$) & 0.66 & ($2^{12}$, $0.6 \times 10^{-6}$) & ($2^{12}$, $0.3 \times 10^{-5}$) & $2^{19}$ & ($2^6$, 6) & $2^{12}$ & $2^{12}$ \\
    \bottomrule
\end{tabular}
}
\label{tab:compression_params1}
\vspace{-3mm}
\end{table*}

\begin{table}[t]
 \vspace{-2mm}
  \caption{6 CL designed for the key parameter of single-compression-parameter methods, generating 6 DL of 3DGS models each accordingly.}
 
  \renewcommand{\arraystretch}{1}
  \centering
  \vspace{-2mm}
  \resizebox{0.47\textwidth}{!}{
  \begin{tabular}{l!{\vrule}c!{\vrule}cccccc}
    \toprule
    \textbf{3DGS Model} & \textbf{Parameter} & \textbf{DL01} & \textbf{DL02} & \textbf{DL03} & \textbf{DL04} & \textbf{DL05} & \textbf{DL06}
    \\
    \midrule
    \textbf{HAC}\cite{HAC} & lambda  & 0.400 & 0.300 & 0.200 & 0.120 & 0.060 & 0.004
    \\
    \textbf{Scaffold}\cite{Scaffold} & vsize & 0.250 & 0.200 & 0.160 & 0.120 & 0.080 & 0.001
    \\
    \bottomrule

  \end{tabular}}
    \label{tab:compression_params2}
\vspace{-3mm}
\end{table}

\subsubsection*{\bf 3DGS Model Selection} 
The 3DGS framework represents a scene using a set of learnable 3D Gaussians. Each Gaussian primitive is characterized by two distinct attribute categories: geometric parameters that specify the Gaussian's spatial characteristics, including position $\mu \in \mathbb{R}^3$, opacity $\alpha \in \mathbb{R}$, covariance matrix $\Sigma$ via scale$s \in \mathbb{R}^3$ and rotation, and appearance parameters that capture view-dependent colors through spherical harmonics (SH) coefficient representation.

To address the storage bottleneck for practical deployment, recent 3DGS methods increasingly integrate compression into the training process, giving rise to 3DGS generative compression algorithms. Our evaluation dataset is constructed from six representative 3DGS generative compression methods:

\begin{itemize}
\item{Compact-3DGS\cite{Compact-3DGS}:  reduces 3DGS storage overhead through learnable Gaussian masking, replacing color representation with hash-grid neural
networks (controlled by \textit{hashmap}), and residual vector quantization \textit{(codebook\_size, rvq\_num)} for geometric attributes, achieving over 25× compression while maintaining quality.}

\item{LightGS\cite{LightGS}: achieves compression of 3DGS through (1) global significance-based pruning \textit{(prune\_percents)}, (2) distilling high-degree SH coefficients to lower-degree representations using knowledge distillation, (3) applying vector quantization to color attributes with learned codebooks \textit{(codebook\_size, vq\_ratio)}.}

\item{CompGS\cite{CompGS}: applies quantization-aware training with K-means clustering, separately quantizing geometry and color attributes into distinct codebooks \textit{(codebook\_size)}, while using run-length encoding for sorted indices.}

\item{C3dGS\cite{c3dgs}: adopts sensitivity-aware vector clustering, where Gaussians exceeding sensitivity thresholds (importance\_include) are preserved while remaining color and geometry parameters are quantized into learned codebooks \textit{(codebook\_size)} via weighted k-means, followed by quantization-aware fine-tuning and entropy encoding.}

\item{HAC\cite{HAC}: exploits spatial consistency among unorganized anchors via a structured binary hash grid, where interpolated hash features predict Gaussian parameters for entropy coding, with the trade-off parameter \textit{lambda} balancing rendering fidelity and entropy loss.}

\item{Scaffold\cite{Scaffold}: restructures 3DGS by using anchor points to spawn and predict local neural Gaussians dynamically, where the \textit{vsize} parameter controls the spatial resolution of the sparse voxel grid that defines anchor placement.}
\end{itemize}

\subsubsection*{\bf Compression Parameter Level Design}  
Investigating their strategies and parameter configurations, we observe that distortion artifacts primarily manifest in two forms: geometric and color distortion. Based on the compression architectures and parameter types, we partition the selected 3DGS methods into two categories: 
(1) \textit{Single-compression-parameter algorithms} utilize anchor-based architectures where a single dominant parameter governs compression behavior, predominantly influencing geometric distortion, including HAC and Scaffold.
(2) \textit{Multi-compression-parameter algorithms} apply specialized compression strategies to different 3DGS attributes through multiple parameters, thereby affecting distortions in both geometric and color aspects, including LightGS, C3dGS, Compact-3DGS, and CompGS.
To enable controlled analysis of distortions introduced by various compression parameters and algorithms, we define differentiated \textit{Compression Levels (CL)} for individual parameters that influence geometry or color quality. Systematic pairwise combinations of these CLs produce diverse 3DGS \textit{Distortion Levels (DL)} with distinct distortion characteristics.

For multi-compression-parameter methods, we configure $\mathbf{4}$ CL respectively for geometry-controlling and color-controlling parameters, as shown in Table~\ref{tab:compression_params1}, creating $\mathbf{4 \times 4 = 16}$ DL through all possible pairings.
For single-compression-parameter methods, we specify $\mathbf{6}$ CL for their dominant compression parameter as shown in Table~\ref{tab:compression_params2}, directly corresponding to $\mathbf{6}$ DL that span a comprehensive quality range.
Our complete dataset comprises $\mathbf{10\ \text{scenes} \times (4 \times 16 + 2 \times 6)\ \text{DLs} = 760}$ trained 3DGS models, which yield $\mathbf{760\ \text{3DGS} \times 20\ \text{viewpoints} = 15,\!200}$ distorted 3DGS rendering images.

\subsection{Subjective Experiment and Data Processing}
\label{Subjective_Experiment_and_Data_Processing}
\subsubsection*{\bf Subjective Experiment Setup and Procedure} 

\begin{figure*}[!t]
    \centering
    \vspace{-5mm}
    \includegraphics[width=1\linewidth]{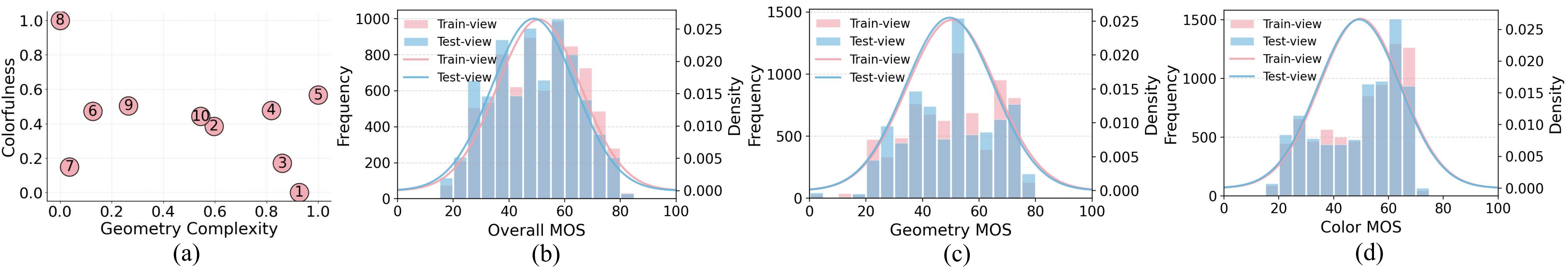}
    \vspace{-8mm}
    \caption{Statistical analysis of 3DGS-IEval-15K+ dataset from content complexity and quality perspectives. (a) Scene distribution across colorfulness and geometry complexity dimensions. (b)-(d) MOS distribution for overall quality, geometric quality, and color quality across training and test viewpoints.} 
    \label{fig:dataset_analyze}
    \vspace{-3mm}
\end{figure*}

\begin{table}[t]
 \vspace{-2mm}
  \caption{Inter-dimensional correlation for 15200 images in 3DGS-IEval-15K+, where subscripts O, G, and C denote Overall, Geometry, and Color quality, respectively.}
 
  \renewcommand{\arraystretch}{1}
  \centering
  \vspace{-2mm}
  \resizebox{0.47\textwidth}{!}{
  \begin{tabular}{c!{\vrule}ccc}
    \toprule
    Metric & $\mathrm{MOS_{O} \to  MOS_{G}}$ & 
    $\mathrm{MOS_{O} \to MOS_{C}}$ & 
    $\mathrm{MOS_{G} \to MOS_{C}}$
    \\
    \midrule
    SRCC & 0.7576  & 0.7856 & 0.6505 \\
    PLCC & 0.7386 & 0.7651 & 0.6136 \\
    KRCC & 0.5657 & 0.5818 & 0.4700\\
    \bottomrule
  \end{tabular}}
    \label{tab:dimension_correlation}
\vspace{-3mm}
\end{table}

To enable fine-grained quality assessment that can decouple and diagnose the dimension-specific distortions derived from 3DGS generative compression, we propose a three-dimensional evaluation framework: \textit{(1) Overall Quality:} assessing general visual fidelity; \textit{(2) Geometry Quality:} evaluating structural integrity, spatial accuracy, and edge preservation; \textit{(3) Color Quality:} measuring color accuracy, saturation, and chromatic consistency. 
Subjects rate each quality dimension using the 11-point impairment scale recommended by ITU-T P.910 \cite{ITU1(siti)}.
We employ a dual-stimulus impairment methodology where the ground-truth reference and its corresponding 3DGS-reconstructed images are presented side-by-side. 
Image presentation and score collection are facilitated through a custom-developed interface implemented using Python Tkinter framework, as illustrated in Figure~\ref{fig:teasor2}(d). 
The experiments are conducted on a calibrated 27-inch AOC Q2790PQ monitor within a controlled indoor laboratory setting, maintaining consistent standard illumination conditions throughout all evaluation sessions. 
To mitigate observer fatigue effects and maintain evaluation accuracy, the complete image collection of 15,200 samples is randomly partitioned into 8 balanced subsets. 
Finally, we obtain a total of 684,000 human annotations (15 valid annotators per iamge
$\times$ 3 dimensions $\times$ 15,200 images).


\subsubsection*{\bf Subjective Data Processing} 

To derive the MOS values for each image, firstly, the raw subjective scores are converted to Z-scores, and then linearly scaled to the [0, 100] range, as demonstrated below:

\begin{equation}
z_{ij}^{} =\frac{r_{ij} - u_{ij}  }{\sigma _{i} } ,   z_{ij}^{'} = \frac{100(z_{ij}+3) }{6} 
\end{equation}

\begin{equation}
\mu _{i}=\frac{1}{N_{i} }\sum_{j=1}^{N_{i} }  r_{ij},  \sigma _{i}=\sqrt{\frac{1}{N_{i}-1 }\sum_{j=1}^{N_{i} }(r_{ij} - \mu _{i})^{2}     },
\end{equation}
where $r_{ij}$ is the raw rating score given by the $i$-th subject to the $j$-th image. $N_{i}$ is the total number of images scored by subject $i$. Next, the MOS of the $j$-th image is derived by computing the average of the rescaled z-scores across all subjects as follows:
\vspace{-1mm}
\begin{equation}
MOS_{j}=\frac{1}{M}\sum_{i=1}^{M}z_{ij}^{'}, 
\end{equation}
where $MOS_{j}$ represents the MOS for the  j-th image, M indicates the total number of subjects, and $z_{ij}^{'}$ refers to the rescaled z-scores. Therefore, a total of 45,600 MOSs (derived from 3  dimensions $\times$ 15,200 images) are obtained.

\subsection{Subjective Data Analysis} 
\subsubsection*{\bf Scene Content Diversity Analysis}

To validate the diversity of our selected original scenes, we assess their colorfulness and geometric complexity through colorfulness metrics (CM) \cite{ITU2} and spatial information (SI) \cite{ITU1(siti)}, respectively. 
Figure ~\ref{fig:dataset_analyze} (a) shows the distribution of our 10 scenes across the colorfulness-geometry complexity space, where each numbered point corresponds to the scene index in Figure~\ref{figure:model}. 
The balanced distribution across both complexity dimensions confirms the representativeness of our scene selection, ensuring that our dataset provides diverse visual challenges and enhancing the generalization of our dataset for quality assessment model training.

\subsubsection*{\bf MOS Analysis}
Figures~\ref{fig:dataset_analyze}(b)-(d) illustrate the MOS distributions for overall, geometric, and color quality respectively, across both training and testing (novel) viewpoints. All three perceptual dimensions exhibit broad quality coverage, which ensures adequate representation of diverse distortion levels and provides balanced training data for quality assessment models. Notably, color quality MOS demonstrates a pronounced rightward skew with elevated sample density in the high-score region ($>$60). This asymmetry arises because the anchor-based compression methods (HAC and Scaffold) predominantly affect geometric quality under compression. Furthermore, for inter-view quality disparity, training viewpoints consistently demonstrate rightward-shifted MOS distributions compared to novel viewpoints across all three dimensions, with higher peaks and mean values. This empirically validates the inherent quality gap between training and novel views in 3DGS. The observed disparity remains modest due to dense training view configurations used in our project. This effect will be substantially more pronounced in sparse-view scenarios.
To evaluate the correlation between different quality dimensions across 15,200 images, we employ three evaluation criteria: Spearman Rank Correlation Coefficient (SRCC), Pearson Linear Correlation Coefficient (PLCC), and Kendall Rank Correlation Coefficient (KRCC). As shown in Table~\ref{tab:dimension_correlation}, the moderate correlations between overall quality and individual dimensions (SRCC: 0.76 for geometry, 0.79 for color) indicate that both dimensions contribute to overall perception but neither dominates. Notably, the lowest correlation appears between geometry and color dimensions, confirming that these two quality aspects are relatively independent and validating the necessity of dimension-specific assessment. 

\begin{figure*}[!t]
    \centering
    \vspace{-2mm}
    \includegraphics[width=1\linewidth]{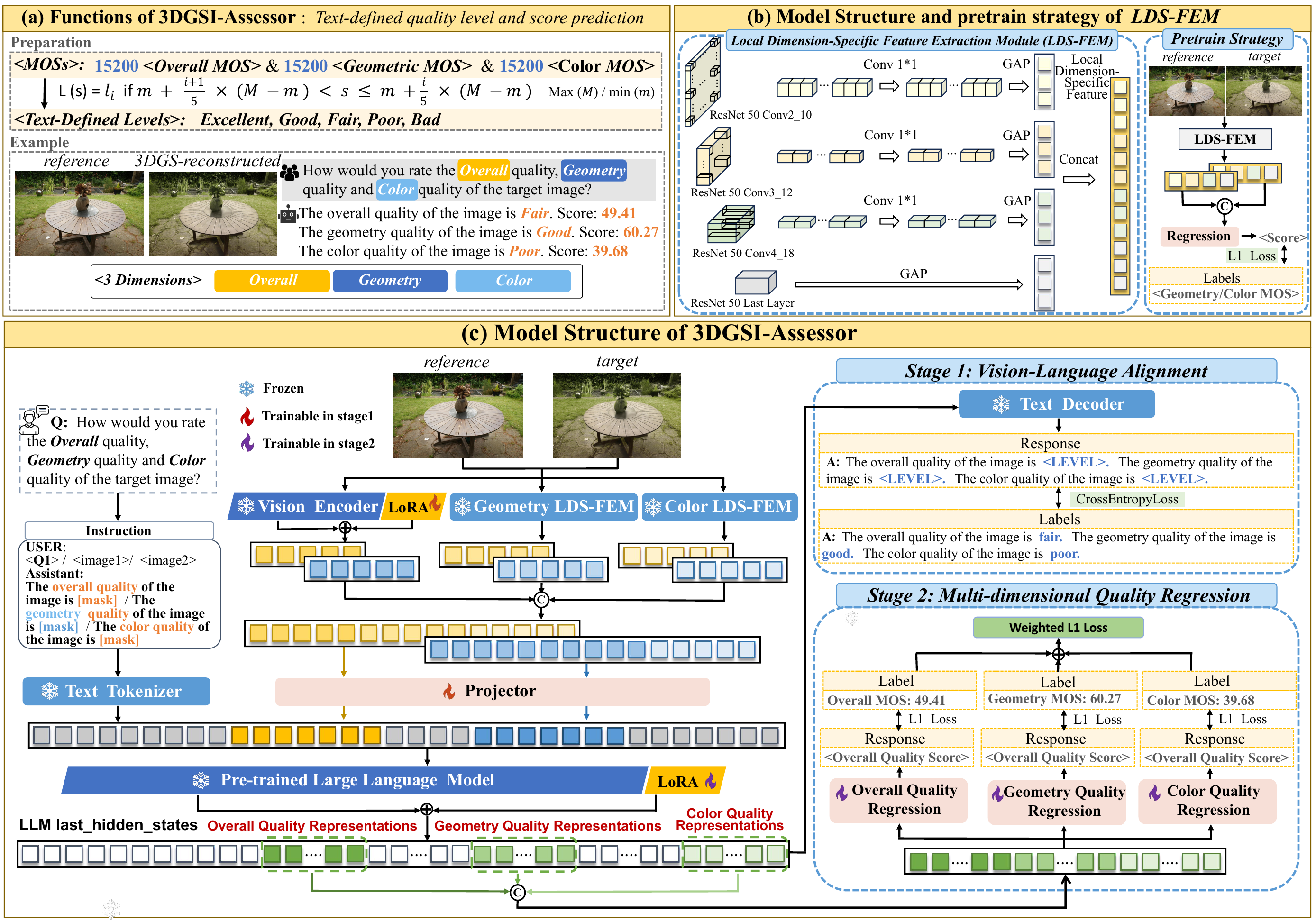} 
    \vspace{-7mm}
    \caption{Framework of 3DGSI-Assessor, an LMM-based method for all-in-one 3DGS image quality assessment. (a) Main functions of 3DGSI-Assessor. The model takes a 3DGS-rendered target image and its reference as inputs and outputs both text-defined quality levels and numerical quality scores across three dimensions (overall, geometry, color). (b) Structure and pretraining strategy of the Local Dimension-Specific Feature Extraction Module (LDS-FEM). Dual LDS-FEM instances are pretrained with geometry and color MOS respectively, extracting multi-scale dimension-specific distortion features from ResNet50 layers. (c) Complete model pipeline demonstrating hierarchical visual encoding (ViT + dual LDS-FEM), visual-linguistic projection, LMM-based multi-modal fusion, and dimension-specific regression heads. Training adopts a two-stage strategy: Stage 1 aligns visual features with the LMM's linguistic space through instruction tuning using text-defined quality levels; Stage 2 specializes the LMM for multi-dimensional score regression.}
    \label{fig:method}
    \vspace{-2mm}
\end{figure*}

\section{The 3DGSI-Assessor Approach}

In this section, we introduce 3DGSI-Assessor, a hierarchical 
LMM-based framework for \textit{all-in-one} 3DGS image quality assessment to provide text-defined quality levels and numerical quality scores for overall, geometry and color dimensions, as shown in Figure~\ref{fig:method}(a).

\subsection{Model Structure}

\subsubsection*{\bf Hierarchical Visual Encoding}

Effective 3DGS quality assessment demands both global semantic understanding and low-level perceptual sensitivity for 3DGS-specific distortions and fine-grained dimensional diagnosis. Therefore, we adopt a hierarchical visual encoding strategy:\\
\textit{(1) Global Semantic Feature Extraction.} We utilize a pre-trained vision transformer (ViT), \textit{i.e.}, InternViT\cite{InternVLmodel} as the global semantic encoder $\mathcal{E}_I$ for holistic quality reasoning. Given the target image $\mathbf{I}_t$ and reference image $\mathbf{I}_r$, the semantic features are extracted as: $\mathbf{F}_I^t = \mathcal{E}_I(\mathbf{I}_t), \quad \mathbf{F}_I^r = \mathcal{E}_I(\mathbf{I}_r)$.
\textit{(2) Local Dimension-Specific Feature Extraction.} To capture low-level perceptual distortions, we introduce the Local Dimension-Specific Feature Extraction Module (LDS-FEM). Specifically, we employ two independent instances of the LDS-FEM, denoted as $\mathcal{E}_G$ and $\mathcal{E}_C$ to extract features for geometry and color dimensions, respectively. As shown in Figure~\ref{fig:method}(b), both extractors are built upon ResNet50 \cite{resnet50} backbones and extract multi-scale features from three intermediate convolutional layers (conv2\_10, conv3\_12, conv4\_18) at different spatial resolutions and receptive fields, enabling detection of dimension-specific degradations manifesting at various scales. For each intermediate layer, we apply channel reduction, adaptive spatial pooling, and feature flattening. These multi-scale features are concatenated with global average pooled features from the final layer, forming compact feature representations. 

Both geometry and color LDS-FEM process the target and reference images simultaneously, obtaining corresponding geometry and color features~($\mathbf{F}^k_i = \mathcal{E}_i(\mathbf{I}_k),~k \in \{t, r\},~i \in \{G, C\}$, where $t = \text{target},~r = \text{reference},~G = \text{geometry},~C = \text{color}$). These features of target and reference images will then be concatenated ($\mathbf{F}^t_G \oplus \mathbf{F}^r _G, \mathbf{F}^t_C \oplus \mathbf{F}^r _C$) and used to regress geometry/color MOSs respectively via $\mathcal{L}_1$ loss on our collected quality datasets. After pre-training, we remove the regression heads and freeze each LDS-FEM as feature encoders. These two encoders then provide stable distortion-aware features inherently aligned with dimension-specific quality attributes, preventing optimization conflicts during LMM fine-tuning. 



\textit{(3) Feature Projection.} To integrate these multi-level representations and align them with the input space of the LMM, we fuse the global semantic features with the local dimension-specific features from LDS-FEM through concatenation and apply a projector $\mathcal{P}_I$ consisting of two multilayer perceptron (MLP) layers to map the visual features into the textual embedding space. The full process can be formulated as:
\begin{equation}
\mathbf{T}_i = \mathcal{P}_I(\mathbf{F}_I^i \oplus \mathbf{F}_G^i \oplus \mathbf{F}_C^i),
\end{equation}
where $i \in \{t, r\}$ denotes target or reference image, and $\mathbf{T}_i$ represents the projected visual tokens.

\subsubsection*{\bf Multi-modal Feature Fusion and Reasoning}

The aforementioned visual features, combined with text embeddings, are fed into a pre-trained LMM (InternVL2.5-8B\cite{InternVLmodel}) for multimodal feature fusion and reasoning to support downstream tasks: (1) quality level descriptions, and (2) quality score prediction.
Concretely, the projected visual tokens $\mathbf{T}_t$ 
and $\mathbf{T}_r$ are injected into LMM at positions corresponding to 
\texttt{<image>} placeholders in the quality assessment prompt:
\begin{quote}
\textit{``Target image: \texttt{<image>} Reference image: \texttt{<image>} How would you rate the overall quality, geometry quality, and color quality 
of the target image?''}
\end{quote}
The LMM processes this multi-modal sequence to jointly model all three quality dimensions, capturing the cross-dimensional dependencies inherent in 3DGS compression. 
Leveraging its superior textual understanding, the model generates textual quality descriptions for each dimension (\textit{e.g.}, ``excellent'', ``good'', ``fair''). This preliminary classification subsequently guides the numerical regression.

\subsubsection*{\bf Multi-dimensional Quality Regression}


We extract hidden states from the LMM's final transformer layer at specific token positions corresponding to quality-level predictions: $\mathbf{h}_{\text{overall}}$, $\mathbf{h}_{\text{geo}}$, 
$\mathbf{h}_{\text{color}}$. These quality-aware features are concatenated 
as $\mathbf{h}_{\text{fusion}} = \mathbf{h}_{\text{overall}} \oplus 
\mathbf{h}_{\text{geo}} \oplus \mathbf{h}_{\text{color}}$ and fed to three 
independent 5-layer MLP regression heads for multi-dimensional MOS prediction:
\begin{equation}
\hat{q}_i = \psi_i(\mathbf{h}_{\text{fusion}}), \quad i \in \{\text{overall}, 
\text{geo}, \text{color}\}.
\end{equation}
Employing independent heads enables dimension-specific optimization, while the shared fusion features preserve cross-dimensional correlations.

\subsection{Training Strategy}

Our training strategy adopts a two-stage curriculum: Stage 1 focuses on vision-language alignment by training the model to generate textual quality descriptions, while Stage 2 specializes the aligned model for multi-dimensional quality regression.

\subsubsection*{\bf Stage 1: Vision-Language Alignment}

Stage 1 aligns the enriched multi-level visual features  with the LMM's linguistic space through textual quality description instruction tuning. Meanwhile, textual quality categorization provides a more interpretable learning signal compared to direct numerical regression, as LMMs are inherently optimized for language modeling rather than continuous value prediction.

We first map continuous MOS scores to five categorical quality levels through uniform interval division. Given the dataset's maximum score $M$ and minimum score $m$, we partition the range $[m, M]$ into five equal intervals: \begin{equation} L(s) = l_i \text{ if } m + \frac{i-1}{5}(M-m) < s \leq m + \frac{i}{5}(M-m), \end{equation} where $\{l_i\}_{i=1}^5 = \{$bad, poor, fair, good, excellent$\}$ follow the ITU\cite{ITU3}.

Given image pair $(\mathbf{I}_t, \mathbf{I}_r)$ and corresponding quality description $\mathbf{y} = [y_1, \ldots, y_L]$ instantiated from the template ``The overall quality is [level$_1$]. The geometry quality is [level$_2$]. The color quality is [level$_3$],'' we optimize the autoregressive language modeling objective:
\begin{equation}
\mathcal{L}_{\text{Stage1}} = -\frac{1}{L}\sum_{i=1}^{L} \log P(y_i \mid 
y_{<i}, \mathbf{I}_t, \mathbf{I}_r; \theta).
\end{equation}
To enhance fine-tuning efficiency, we adopt the LoRA\cite{LORA} technique, which enables parameter-efficient fine-tuning by updating only low-rank decomposition matrices rather than the full model weights. Specifically, we apply LoRA adapters to the ViT encoder $\mathcal{E}_I$ and fully train the projector $\mathcal{P}_I$ from 
scratch, while freezing the pre-trained LDS-FEM ($\mathcal{E}_G$, $\mathcal{E}_C$) and LMM. This configuration preserves the LMM's linguistic capabilities while establishing robust vision-language alignment through visual-side adaptation.

\subsubsection*{\bf Stage 2: Multi-dimensional Quality Regression}

Building upon the aligned model from Stage 1, Stage 2 introduces the three dimension-specific 5-layer MLP regression heads for end-to-end quality score prediction. We employ LoRA to adapt the LMM , and fully train three regression heads $\{\psi_{\text{overall}}, \psi_{\text{geo}}, \psi_{\text{color}}\}$ from random initialization, while freezing all visual components trained in Stage 1.

Given ground-truth quality scores $\{q_{\text{overall}}, q_{\text{geo}}, q_{\text{color}}\}$ and predictions $\{\hat{q}_{\text{overall}}, \hat{q}_{\text{geo}}, \hat{q}_{\text{color}}\}$, we employ multi-task $\mathcal{L}_1$ regression loss:
\begin{equation}
\mathcal{L}_{\text{Stage2}} =\lambda_{1}|\hat{q}_{\text{overall}} - q_{\text{overall}}| + \lambda_{2}|\hat{q}_{\text{geo}} - q_{\text{geo}}| + \lambda_{3}|\hat{q}_{\text{color}} - q_{\text{color}}|.
\end{equation}
where $\lambda_1$, $\lambda_2$, and $\lambda_3$ are penalty weights for overall, geometry, and color quality prediction errors, respectively. These weights can be adjusted to emphasize specific dimensions for different application scenarios. In this work, we set $\lambda_1 = \lambda_2 =\lambda_3 = 1$ for balanced multi-dimensional optimization.
This end-to-end training strategy enables the LMM to adapt its internal representations for precise quality score prediction while preserving the established vision-language alignment.
\section{Experiments}
\begin{table*}[t]
\caption{Performance benchmark on 3DGS-IEval-15K+. $\spadesuit$ Traditional Handcrafted QA models, $\heartsuit$ LMM-based models, $\blacklozenge$ Deep learning-based IQA models. Best results are marked in {\red{RED}} and second-best in {\blue{BLUE}}. *Refers to finetuned models.}
\renewcommand{\arraystretch}{1}
\centering
\vspace{-2mm}
\resizebox{0.9\textwidth}{!}{
  \begin{tabular}{l||ccc|ccc|ccc}
    \toprule
    \multicolumn{1}{l}{\bf Distortion Type} &
    \multicolumn{3}{c}{\textbf{Overall Quality}} &
    \multicolumn{3}{c}{\textbf{Geometry Quality}} &
    \multicolumn{3}{c}{\textbf{Color Quality}} \\
    \cmidrule(lr){2-4} \cmidrule(lr){5-7} \cmidrule(lr){8-10} 
    \textbf{Methods / Metrics} & \textbf{SRCC} & \textbf{PLCC} & \textbf{KRCC} & \textbf{SRCC} & \textbf{PLCC} & \textbf{KRCC} & \textbf{SRCC} & \textbf{PLCC} & \textbf{KRCC} \\
    \hline
    $\spadesuit$ PSNR\cite{PSNR} & 0.6460 & 0.6395 & 0.4571 & 0.6031 & 0.5872 & 0.4291 & 0.5873 & 0.5624 & 0.4102 \\
    $\spadesuit$ SSIM\cite{PSNR} & 0.6797 & 0.6656 & 0.4884 & 0.6784 & 0.6467 & 0.4933 & 0.5666 & 0.5313 & 0.3954 \\
    $\spadesuit$ MS-SSIM\cite{MS-SSIM}& 0.6988 & 0.6773 & 0.5054 & 0.6889 & 0.6483 & 0.5024 & 0.5786 & 0.5358 & 0.4046  \\
    $\spadesuit$ IW-SSIM\cite{IW-SSIM} & 0.6947 & 0.6919 & 0.5052 & 0.7249 & 0.6943 & 0.5372 & 0.5343 & 0.5017 & 0.3718
 \\
    $\spadesuit$ VIF\cite{VIF} & 0.5589 & 0.5689 & 0.3918 & 0.5652 & 0.5569 & 0.4009 & 0.4056 & 0.3925 & 0.2774 \\
    $\spadesuit$ FSIM\cite{FSIM} & 0.7333 & 0.7128 & 0.5355 & 0.7291 & 0.6892 & 0.5388 & 0.5911 & 0.5465 & 0.4130 \\
    $\spadesuit$ BRISQUE\cite{BRISQUE} & 0.2205 & 0.2255 & 0.1493 & 0.2943 & 0.2814 & 0.2031 & 0.1364 & 0.1234 & 0.0924 \\
    $\spadesuit$ LPIPS\cite{LPIPS} & 0.6767 & 0.6679 & 0.4870 & 0.6882 & 0.6579 & 0.5014 & 0.5179 & 0.4606 & 0.3568 \\
    $\spadesuit$ DISTS \cite{DISTS} & 0.8198 & 0.8132 & 0.6215 & 0.7813 & 0.7511 & 0.5875 & 0.6806 & 0.6513 & 0.4855 \\
    \midrule
    $\heartsuit$ LLaVA-one-vision (0.5B)\cite{Llava-one-vision} & 0.2267 & 0.2298 & 0.1825 & 0.1325 & 0.1228 & 0.0652 & 0.1256 & 0.1238 & 0.0728 \\
    $\heartsuit$ LLaVA-one-vision (7B)\cite{Llava-one-vision} & 0.4637 & 0.4865 & 0.3796 & 0.3687 & 0.3613 & 0.2817 & 0.3725 & 0.3612 & 0.2659 \\
    $\heartsuit$ DeepSeekVL (7B)\cite{DeepSeekVL} & 0.7037 & 0.6862 & 0.5456 & 0.6230 & 0.5834 & 0.4834 & 0.3558 & 0.3510 & 0.2638 \\
    $\heartsuit$ LLaVA-1.5 (7B)\cite{LLaVA-1.5} & 0.5723 & 0.5561 & 0.4486 & 0.4561 & 0.4697 & 0.3726 & 0.4659 & 0.4678 & 0.3627 \\
    $\heartsuit$ LLaVA-NeXT (8B)\cite{LLaVA-NeXT} & 0.5886 & 0.5867 & 0.4782 & 0.5929 & 0.5686 & 0.4835 & 0.5851 & 0.5643 & 0.4636 \\
    $\heartsuit$ mPLUG-Owl3 (7B)\cite{mPLUG-Owl3}& 0.3582 & 0.3536 & 0.2665 & 0.2902 & 0.2686 & 0.2199 & 0.3107 & 0.2789 & 0.2291  \\
    $\heartsuit$ CLIPIQA\cite{CLIPIQA} & 0.3629 & 0.3362 & 0.2456 & 0.3705 & 0.3370 & 0.2519 & 0.2702 & 0.2121 & 0.1855 \\
    $\heartsuit$ Qwen2.5-VL (7B)\cite{Qwen2.5-VL} & 0.7265 & 0.6976 & 0.5816 & 0.6606 & 0.6091 & 0.5198 & 0.6342 & 0.6043 & 0.4977  \\
    $\heartsuit$ CogAgent (18B)\cite{CogAgent} & 0.5249 & 0.5116 & 0.4062 & 0.4134 & 0.3750 & 0.3305 & 0.2183 & 0.1546 & 0.1673  \\
    $\heartsuit$ InternVL2.5 (8B)\cite{InternVLmodel} & 0.6678 & 0.6921 & 0.5302 & 0.5980 & 0.5380 & 0.4759 & 0.6106 & 0.5529 & 0.4782  \\
    $\heartsuit$ InternVL3 (9B)\cite{InternVLmodel} & 0.5067 & 0.5316 & 0.3776 & 0.3951 & 0.3950 & 0.2986 & 0.2212 & 0.2423 & 0.1554 \\
    
    $\heartsuit$ Gemini1.5-pro\cite{Gemini1.5} & 0.6872 & 0.6872 & 0.3776 & 0.5427 & 0.5162 & 0.4029 & 0.5728 & 0.5668 & 0.4656  \\
    $\heartsuit$ Q-Align\cite{Q-Align}
    & 0.7718 & 0.7654 & 0.5660 & 0.7550 & 0.7268 & 0.5555 & 0.5994 & 0.5542 & 0.4124 \\
    $\heartsuit$ Qwen2.5-VL (7B)* \cite{Qwen2.5-VL} & 0.9023 & 0.9031 & 0.7212 & 0.8127 & 0.7973 & 0.6413 & 0.8804 & 0.9076 & 0.7145  \\
    $\heartsuit$ Llama3.2-Vision (11B)* \cite{llama3.2} & 0.8922 & 0.8957 & 0.7069 & 0.8268 & 0.8129 & 0.6631 & 0.8716 & 0.8826 & 0.6977  \\
    \midrule
    $\blacklozenge$ TReS\cite{TReS} 
    & 0.8115 & 0.8067 & 0.6131 & 0.8110 & 0.8062 & 0.6125 & 0.8409 & 0.8375 & 0.6475 \\
    $\blacklozenge$ DBCNN\cite{DBCNN}  
    & 0.9202 & 0.9177 & 0.7552 & 0.8418 & 0.8226 & 0.6576 & 0.9125 & 0.9299 & 0.7464 \\
    $\blacklozenge$ MANIQA\cite{MANIQA} 
    & 0.9298 & 0.9269 & 0.7715 & 0.8532 & 0.8373 & 0.6728 & 0.9249 & \blue{0.9417} & 0.7687 \\
    $\blacklozenge$ STAIRIQA\cite{STAIRIQA} 
    & 0.9385 & 0.9365 & 0.7810 & \blue{0.8600} & \blue{0.8414} & \blue{0.6785} & 0.9210 & 0.9384 & \blue{0.7708} \\
    $\blacklozenge$ MUSIQ\cite{MUSIQ} 
    & 0.9082 & 0.9049 & 0.7295 & 0.8359 & 0.8176 & 0.6480 & 0.9039 & 0.9204 &  0.7252  \\
    $\blacklozenge$ HYPERIQA\cite{HYPERIQA} 
    & \blue{0.9406} & \blue{0.9377} & \blue{0.8026} & 0.8368 & 0.8249 & 0.6632 & \blue{0.9269} & 0.9286 & 0.7459   \\
    $\blacklozenge$ LIQE\cite{LIQE} 
    & 0.9274 & 0.9111 & 0.7620 & 0.8499 & 0.8267 & 0.6668 & 0.9261 & 0.9254 & 0.7519  \\
    \midrule
\rowcolor{gray!20}  \textbf{3DGSI-Assessor (Ours)}
    & \red{0.9630} & \red{0.9663} & \red{0.8482} & \red{0.9489} & \red{0.9634} & \red{0.8214} & \red{0.9601} & \red{0.9732} & \red{0.8468}  \\

    \bottomrule
  \end{tabular}
}
\label{tab:benchmark}
\vspace{-1mm}
\end{table*}

\subsection{Experiment Setup}
\subsubsection*{\bf Evaluation Datasets and Metrics}Our proposed method is validated on four NVS QA datasets: 3DGS-IEval-15K+, ENeRF-QA\cite{ENeRF-QA}, NeRF-VSQA\cite{NeRF-VSQA}, and GSC-QA\cite{GSC-QA}. 
To evaluate the correlation between the predicted scores and the ground-truth MOSs, we employ three evaluation criteria: SRCC, PLCC, and KRCC.

\subsubsection*{\bf Implementation Details for QA Models} To assess the performance of
our proposed method, we select SOTA QA metrics for comparison, which can be classified into three groups: 
\begin{itemize}
\item{Traditional Handcrafted QA Metrics: PSNR\cite{PSNR}, SSIM\cite{PSNR}, LPIPS\cite{LPIPS}, VIF\cite{VIF}, \textit{etc.} Without training, these methods are directly applied to the corresponding databases, producing a single quality score per image.} 
\item{LMM-based Models: DeepSeekVL\cite{DeepSeekVL}, LLaVA-1.5 \cite{LLaVA-1.5}, mPLUG-Owl3 \cite{mPLUG-Owl3}, Qwen2.5-VL\cite{Qwen2.5-VL}, \textit{etc.} 
For these models, we perform a fair zero-shot evaluation utilizing their publicly available pre-trained weights for inference without any dataset-specific fine-tuning. Using the same training and testing split as our model, we further fine-tune two of these LMMs using the same LLM fine-tuning approach in our model for rigorous and fair comparison. To construct a more stringent baseline, we perform dimension-wise training and inference separately for each of the three quality dimensions, thereby testing whether our unified all-in-one approach can outperform dimension-specific LMM baselines.}  
\item{Deep Learning-based Models: DBCNN\cite{DBCNN}, HYPERIQA\cite{HYPERIQA}, MANIQA\cite{MANIQA}, MUSIQ\cite{MUSIQ}, \textit{etc.} 
For these models, we adhere to the protocols from their original publications and maintain the identical 4:1 training and testing split used for our model. Since these architectures are designed to learn a single quality dimension, we train models independently for each of the three dimensions.}  
\end{itemize}

Our proposed model is implemented with PyTorch and trained on a 48\,GB NVIDIA GeForce RTX 4090 GPU with a batch size of 8. The initial learning rate is set to $1\mathrm{e}{-5}$ and decayed with a cosine annealing schedule. We adopt the Adam optimizer with $\beta_{1}=0.9$ and $\beta_{2}=0.999$, and train for 5 epochs in each of the vision-language alignment and quality-regression stages. 
Unless otherwise stated, we adopt a within-scene split: for each scene, the rendered images are partitioned 4:1 into training and testing sets, such that no image appears in both, and results are averaged over 5-fold cross-validation. 
At inference, 3DGSI-Assessor requires about 18\,GB of GPU memory and processes each image in roughly 0.11\,s on a single RTX 4090, while jointly producing all three quality dimensions in one forward pass. This cost is well within the budget of practical 3DGS quality assessment and runs on a single commodity GPU. Notably, the method has been adopted as a baseline in the \emph{IEEE Standards Association project 3366.5, Standard for Gaussian Splats Quality}, evidencing its practical value for real-world 3DGS quality evaluation.


\begin{table*}[t]
  \centering
    \caption{Zero-shot cross-dataset performance on multiple benchmarks. Best results are marked in {\red{RED}} and second-best in {\blue{BLUE}}. } 
    \vspace{-1mm}
  \label{tab:method}%
 
  \renewcommand\arraystretch{1}
  \resizebox{0.85\textwidth}{!}{
    \begin{tabular}{l|ccc|ccc|ccc}
    \Xhline{1px}
    \multirow{2}[1]{*}{Method} & \multicolumn{3}{c|}{ENeRF-QA\cite{ENeRF-QA}} & \multicolumn{3}{c|}{NeRF-VSQA\cite{NeRF-VSQA}}  & \multicolumn{3}{c}{GSC-QA\cite{GSC-QA}} \\
\cmidrule{2-10}    & SRCC &PLCC & KRCC  & SRCC &PLCC & KRCC  & SRCC &PLCC & KRCC\\
    \midrule
    TReS\cite{TReS} & 0.7526 & 0.7165  & 0.5467 
    & 0.5016 & 0.6107 & 0.3769 
    & 0.5777 & 0.5516 & 0.4214 \\
    DBCNN\cite{DBCNN}  
    & 0.6598 & 0.6726  & 0.4812 
    & 0.5232 & 0.6240 & 0.3804
    & \blue{0.7697} & 0.7647 & \blue{0.5764} \\
    HYPERIQA\cite{HYPERIQA} 
    & 0.7456 & 0.7529  & 0.5613 
    & 0.5978 & 0.6811 & 0.4462
    & 0.5892 & 0.5859 & 0.4284 \\
    LIQE\cite{LIQE} 
    & 0.7426 & \blue{0.8137}  & 0.5589 
    & 0.6001 & 0.6991 & 0.4504
    & 0.7596 & \blue{0.7695} & 0.5708  \\
    MANIQA\cite{MANIQA}  
    & \blue{0.7828} & 0.7736  & \blue{0.5829}
    & \blue{0.6995} & \blue{0.7172} & \blue{0.5298}
    & 0.7659 & 0.7692 & 0.5762 \\
    MUSIQ\cite{MUSIQ} 
    & 0.6985 & 0.7236  & 0.5122 
    & 0.5246 & 0.6299 & 0.3851 
    & 0.6402 & 0.6232 & 0.4663 \\
    STAIRIQA\cite{STAIRIQA} 
    & 0.6445 & 0.6626 & 0.4637
    & 0.5320 & 0.6278 & 0.3861
    & 0.6081 & 0.5621 & 0.4559 \\
     \midrule
    \rowcolor{mycolor_gray} \textbf{3DGSI-Assessor (Ours)} 
    & \red{0.8196} & \red{0.8417} &\red{0.6539}
    & \red{0.7669} & \red{0.7754}&\red{0.5756} 
    & \red{0.7769} &\red{0.7769} & \red{0.5800} \\

    \Xhline{1px}
    \end{tabular}%
  }
\label{cross-dataset}
\end{table*}%

\begin{table*}[t]
\renewcommand\arraystretch{1}
  \caption{Ablation study on hierarchical visual encoding , quality-aware vision-language training, and LoRA strategy of 3DGSI-Assessor.}
\vspace{-2mm}

  \resizebox{1\textwidth}{!}{
  \begin{tabular}{ccccccc|ccc:ccc:ccc}
    \Xhline{1px}
& \multicolumn{6}{c}{\textbf{Feature \& Strategy}}  & \multicolumn{3}{c}{\textbf{Overall Quality }} & \multicolumn{3}{c}{\textbf{Geometric Quality }} &  \multicolumn{3}{c}{\textbf{Color Quality}} \\
    \cline{2-16} 
    \multicolumn{1}{c}{No.}   & LLM & VIT  & LDS-FEM & quality & LoRA$_{r=16}$(vit)    & LoRA$_{r=16}$(llm) & SRCC    & PLCC  &KRCC  & SRCC     & PLCC    & KRCC   &    SRCC     & PLCC   & KRCC              \\
    \hline
    \multicolumn{1}{c}{(1)}  &   &   & \ding{52} &   &   &    
    & 0.8962 & 0.9018 & 0.7369 & 0.8302  & 0.8012  &  0.6437  &  0.8999  & 0.9287  &  0.7293  \\
    \multicolumn{1}{c}{(2)}  &   & \ding{52}  &   &   &   \ding{52} &     
    & 0.9232 & 0.9221 &  0.7638  &  0.8419  &  0.8224  & 0.6618  &  0.9194  & 0.9369 & 0.7619  \\
    \multicolumn{1}{c}{(3)}  & \ding{52}  &  \ding{52} &   &  \ding{52} &  \ding{52} &  \ding{52}   
    &  0.9562  &  0.9590  &  0.8395  & 0.9368 & 0.9527 & 0.8090  &  0.9480  &  0.9622 &  0.8381     \\
    \multicolumn{1}{c}{(4)}  &  \ding{52} & \ding{52}  & \ding{52} &   &  \ding{52} & \ding{52}  
    & 0.9578 &  0.9586 & 0.8411 &  0.9425  & 0.9587  & 0.8165 & 0.9526 &  0.9674 &  0.8459    \\
    \multicolumn{1}{c}{(5)}  & \ding{52}  &  \ding{52} & \ding{52} &  \ding{52} &   &   
    & 0.9386 & 0.9377 & 0.8261 & 0.9282 & 0.9376 &  0.7991 & 0.9322 & 0.9416 & 0.8187   \\
    \multicolumn{1}{c}{(6)}  &  \ding{52} & \ding{52}  & \ding{52} & \ding{52}  &   &  \ding{52} 
    &  0.9465  & 0.9472  &  0.8319 & 0.9391 & 0.9553 & 0.8075 & 0.9532 & 0.9668 & 0.8396  \\
    \multicolumn{1}{c}{(7)}  &  \ding{52} & \ding{52}  & \ding{52} &  \ding{52} & \ding{52} &
    & 0.9516  & 0.9578  &  0.8397  &  0.9332  &  0.9527  & 0.8021  &  0.9519 &   0.9654 & 0.8357  \\
    \rowcolor{gray!20} \multicolumn{1}{c}{(8)}          &   \ding{52}      &  \ding{52}         & \ding{52} &      \ding{52}         & \ding{52}    & \ding{52}   
    &  \textbf{0.9630}  &  \textbf{0.9663}  &  \textbf{0.8482}  &  \textbf{0.9489} & \textbf{0.9634}  &   \textbf{0.8214}  & \textbf{0.9601} & \textbf{0.9732} & \textbf{0.8468}   \\
    
    \Xhline{1px}
  \end{tabular}\label{ablation}
   }
  \centering
\end{table*}

\subsection{Evaluation on the 3DGS-IEval-15K+ Database}

As shown in Table~\ref{tab:benchmark}, traditional handcrafted metrics such as PSNR and SSIM \cite{PSNR} correlate weakly with human perception and cannot provide dimension-specific assessment. Zero-shot LMMs vary widely: smaller models struggle across all dimensions, while larger ones such as Qwen2.5-VL \cite{Qwen2.5-VL} improve but remain insufficient, as they emphasize semantic over low-level perceptual features. Fine-tuning these LMMs with the same strategy used in our model improves their performance substantially, confirming the effectiveness of our fine-tuning design. Deep learning-based methods are competitive on overall quality but must be trained independently for each dimension, precluding cross-dimensional reasoning and multiplying training cost. 
In contrast, 3DGSI-Assessor delivers consistent SOTA performance across all three dimensions with a single model. Its advantage is most pronounced on the diagnostic dimensions that matter for compression tuning: it improves SRCC over the strongest baseline by 10.3\% on geometry and 3.6\% on color, while also leading on overall quality. This confirms that the all-in-one design not only unifies multi-dimensional assessment in one model but is especially effective precisely where dimension-specific diagnosis is required.

\subsection{Zero-shot Cross-dataset Evaluation}
To assess generalization, all learning-based baselines and 3DGSI-Assessor are trained on 3DGS-IEval-15K+ and evaluated without fine-tuning on three external benchmarks spanning both NVS representations (NeRF and 3DGS) and both synthetic and real-world scenes: ENeRF-QA \cite{ENeRF-QA}, NeRF-VSQA \cite{NeRF-VSQA}, and GSC-QA \cite{GSC-QA}. 

As shown in Table~\ref{cross-dataset}, 3DGSI-Assessor ranks first on all three benchmarks under every correlation measure. Moreover, among the baselines no method transfers consistently: HYPERIQA, one of the strongest baselines on the in-domain benchmark (Table~\ref{tab:benchmark}), generalizes only moderately here, ranking mid-pack on all three external benchmarks—strong in-domain fit does not imply cross-domain robustness. More broadly, DBCNN is the runner-up on GSC-QA yet drops to $0.6598$ SRCC on ENeRF-QA, whereas MANIQA leads the baselines on ENeRF-QA and NeRF-VSQA but is weaker on GSC-QA. Each transfers well only where the target distortions resemble its training domain, while 3DGSI-Assessor remains the top performer under every condition. This consistency across representations and scene types is difficult to attribute to scene memorization, indicating that 3DGSI-Assessor captures transferable, 3DGS-relevant distortion characteristics.

\subsection{Ablation Study}

We conduct ablation experiments to validate the effectiveness of different components in our 3DGSI-Assessor method, with results summarized in Table \ref{ablation}. Our analysis reveals three key findings: First, experiments (1)-(3), and (8) demonstrate the effectiveness of our hierarchical visual encoding strategy. Experiments (1) and (2), using standalone ViT or LDS-FEM with direct regression, achieve limited performance. Experiment (3) introduces the LLM for multi-modal feature fusion, significantly improving prediction correlation, validating the LLM's capability in bridging visual features and quality reasoning. Experiment (8) further incorporates the pre-trained LDS-FEM, enhancing performance across all dimensions, with particularly notable gains in geometric (+0.6\%) and color (+0.5\%) quality, confirming that local dimension-specific features effectively complement global semantic representations for capturing 3DGS-specific distortions. Second, comparing experiments (4) and (8), we observe that Stage 1 training with quality level description improves performance by aligning visual features with the LMM's linguistic space through instruction-tuning, leveraging the LMM's strength in language understanding for subsequent numerical regression. Third, through experiments (5)-(8), we validate the significant performance gains achieved by LoRA fine-tuning. Experiment (8), which integrates all proposed components (hierarchical visual encoding, the two-stage training strategy, and LoRA fine-tuning), achieves the best performance across all dimensions, demonstrating the necessity of our unified framework for comprehensive 3DGS quality assessment.

\section{Limitations and Future Work}
In this section, we discuss the limitations of the present work and potential directions for future research. First, 3DGS-IEval-15K+ is built on 10 diverse real-world scenes, which constitute the dominant application scenario of 3DGS; extending the dataset with synthetic content in future work would provide even broader coverage of the NVS domain. Second, as the dimension-specific predictions of 3DGSI-Assessor can be used to guide the separate tuning of geometric and color compression parameters, future work can integrate the metric as a perceptual reward inside 3DGS training and rate-distortion optimization.

\section{Conclusion}
In this paper, we introduce 3DGS-IEval-15K+, a multi-dimensional IQA dataset specifically designed for compressed 3DGS, comprising 15,200 images from 10 diverse scenes with 45,600 MOSs across overall, geometry, and color quality dimensions, which is the largest 3DGS IQA dataset to date.
Based on 3DGS-IEval-15K+, we propose 3DGSI-Assessor, the first fine-grained, LMM-based IQA framework tailored for 3DGS that leverages vision-language alignment and LoRA fine-tuning to achieve all-in-one multi-dimensional quality assessment. Extensive experiments demonstrate that 3DGSI-Assessor achieves SOTA performance on 3DGS-IEval-15K+ and exhibits competitive zero-shot generalization ability on other NVS evaluation benchmarks. 
Our work establishes a foundation for 3DGS quality assessment.

\bibliographystyle{IEEEtran}
\bibliography{main.bib}

\vfill

\end{document}